\documentclass[10pt,twocolumn,letterpaper]{article}

\usepackage{cvpr}              % To produce the CAMERA-READY version
\usepackage[dvipsnames]{xcolor}

\definecolor{cvprblue}{rgb}{0.21,0.49,0.74}
\usepackage[pagebackref,breaklinks,colorlinks,citecolor=cvprblue]{hyperref}
\usepackage{algorithm}
\usepackage{xcolor}
\usepackage{algpseudocode}
\usepackage{marvosym}

\usepackage[algo2e,linesnumbered,noend]{algorithm2e}

\usepackage{multirow}

\def\paperID{15747} % *** Enter the Paper ID here
\def\confName{CVPR}
\def\confYear{2024}

\title{Mitigating Noisy Correspondence by Geometrical Structure Consistency Learning}

\author{Zihua Zhao$^1$, Mengxi Chen$^1$, Tianjie Dai$^{1,2}$, Jiangchao Yao$\textsuperscript{1,2\ \Letter}$, Bo Han$^3$, \\ Ya Zhang$^{1,2}$, Yanfeng Wang$\textsuperscript{1,2\ \Letter}$\\
$^1$ Cooperative Medianet Innovation Center, Shanghai Jiao Tong University, \\ $^2$ Shanghai Artificial Intelligence Laboratory, $^3$ Hong Kong Baptist University\\
{\tt\small \{sjtuszzh, mxchen\_mc, elfenreigen, Sunarker, ya\_zhang, wangyanfeng\}@sjtu.edu.cn}, \\ \tt\small bhanml@comp.hkbu.edu.hk
}

\usepackage[accsupp]{axessibility} % Improves PDF readability for those with visual impairments.
\begin{document}
\maketitle
\begin{abstract}

Noisy correspondence that refers to mismatches in cross-modal data pairs, is prevalent on human-annotated or web-crawled datasets. Prior approaches to leverage such data mainly consider the application of uni-modal noisy label learning without amending the impact on both cross-modal and intra-modal geometrical structures in multimodal learning. Actually, we find that both structures are effective to discriminate noisy correspondence through structural differences when being well-established. Inspired by this observation, we introduce a Geometrical Structure Consistency (GSC) method to infer the true correspondence. Specifically, GSC ensures the preservation of geometrical structures within and between modalities, allowing for the accurate discrimination of noisy samples based on structural differences. Utilizing these inferred true correspondence labels, GSC refines the learning of geometrical structures by filtering out the noisy samples. Experiments across four cross-modal datasets confirm that GSC effectively identifies noisy samples and significantly outperforms the current leading methods. Source code is available at: \url{https://github.com/MediaBrain-SJTU/GSC}.

\end{abstract}    
\section{Introduction}
\label{Section: Introduction}

\begin{figure}[t]
\begin{center}
\includegraphics[width=0.45\textwidth]{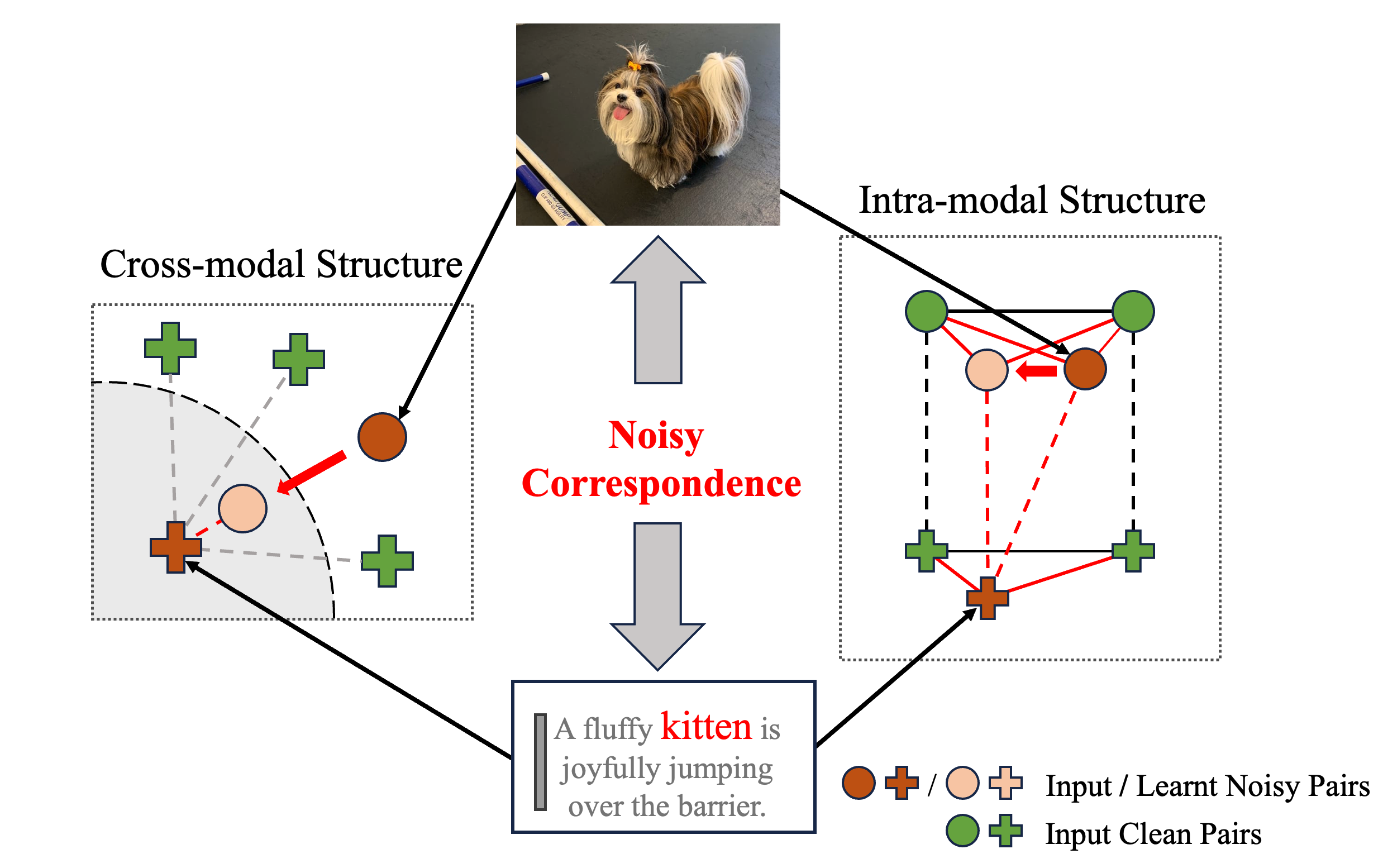}
% [width=1.30\textwidth]
\caption{Noisy correspondence impacts both cross-modal and intra-modal geometrical structures. \textbf{Left:} Cross-modal distance between mismatched text and image is initially distant but wrongly reduced. \textbf{Right:} Intra-modal structures of mismatched image (above) and text (below) are initially distinct but wrongly aligned, thus similar samples within the modality are pulled apart.}
\label{Figure: Introduction}
\end{center}
\vskip -0.3in
\end{figure}

Cross-modal retrieval~\citep{lu2022cots, kim2023exposing, liu2022self} that focuses on querying the most relevant samples across modalities, has garnered considerable interest in multimodal scenarios~\citep{xu2023multimodal, baltruvsaitis2018multimodal}. Most current methods presuppose large quantity of well-annotated data is available. However, real-world datasets~\citep{sharma2018conceptual, jia2021scaling, chua2009nus} which own non-expert annotation or collected by web crawling are prone to noisy correspondence. Such discrepancy between modalities can cause severe degradation to retrieval models if without proper handling~\citep{huang2021learning, qin2022deep}.

Recently, learning with noisy correspondence has gathered increasing attention. The majority of these efforts share a common target of accurately learning the true soft correspondence labels that can reliably indicate the matching degree between data pairs. For example, NCR~\citep{huang2021learning} pioneered this area by employing a co-teaching approach to classify samples with higher losses as noisy. Such method, which has been further refined by~\citep{yang2023bicro, han2023noisy} by introducing meta-learning and leveraging clean data subsets, however fundamentally remains a variation of the uni-modal sample selection philosophy~\citep{li2020dividemix}. They may not be very effective to identify the accurate correspondence and finally overfit on noise in face of the intricacies of multimodal learning. 

% Distinct from the noisy label problem in uni-modal learning, noisy correspondence affects the geometrical structures of multimodal representations from both cross-modal and intra-modal aspects. 

Noisy correspondence affects multimodal representations in a more complicated way than uni-modal noisy label problem.
% by damaging both cross-modal and intra-modal geometrical structures. 
As illustrated in Fig.~\ref{Figure: Introduction}, in the perspective of the cross-modal geometrical structure that refers to the similarities between representations across modalities, the presence of noisy correspondence can disrupt this structure by erroneously reducing the distance between mismatched data pairs. On the other hand, in the perspective of the intra-modal geometrical structure that refers to the similarities within the modality, as different samples exhibit asymmetric intra-modal structures, attempting to align the distinct structures across mismatched modalities can lead to intra-model collapse. Fortunately, we have observed significant differences between clean and noisy samples within well-established cross-modal and intra-modal geometrical structures where clean samples tend to show better alignments, as detailed in Sec.~\ref{Section: Motivation}. These differences conversely offer a promising strategy for discriminating noisy samples and accurately predicting true correspondence labels.

% Distinct from the noisy label problem in uni-modal learning, noisy correspondence affects the geometrical structures of multimodal representations from both cross-modal and intra-modal aspects. Consider the contrastive learning optimization process, as depicted in Fig.~\ref{Figure: Introduction}, the paired text acts as a positive key to the query image, contrasting with negative keys from other texts. If the paired data is noisy, the cross-modal distance between the query and its positive key is erroneously reduced. Concurrently, the query image is drawn closer to a cluster of images that resemble the text of the noisy positive key. Despite these challenges, we have observed significant differences between clean and noisy samples within well-established cross-modal and intra-modal geometrical structures (as detailed in Section~\ref{Section: Motivation}). These differences offer a promising strategy for discriminating noisy samples and accurately predicting true correspondence labels.

Inspired by the distinct structural characteristics, we introduce the Geometrical Structure Consistency (GSC) method to mitigate the issue of noisy correspondence. Specifically, GSC maintains the integrity of geometrical structures by optimizing a contrastive loss that aligns the intra-modal geometrical structures along with the traditional loss for cross-modal alignment. Benefiting from the memorization effect of DNNs~\citep{xia2020robust, arpit2017closer}, geometrical structure can be well-established in the early stage. During this phase, GSC assesses the true soft correspondence labels based on the differences in geometrical structure within and across modalities. For the cross-modal aspect, GSC identifies potential noise by recognizing data pairs with low cross-modal similarities, while in the case of intra-modal aspect, the similarity between the queried sample and other samples are calculated to determine the intra-modal geometrical structure, and data pairs with inconsistent structures across modalities are considered as noisy. These assessed labels are then used to clarify the learning of the geometrical structure, creating a positive feedback loop. To conclude, the contributions of this work are summarized as follows:
\begin{itemize}
    \item We identify the impact of noisy correspondence on both cross-modal and intra-modal geometrical structures, and find the significant difference between clean samples and noisy samples within a well-established overall structure.
    \item We introduce the novel Geometrical Structure Consistency (GSC) approach, which utilizes the structural differences to accurately predict true correspondence labels and counteract the adverse effects of noisy correspondences.
    \item Our proposed GSC method is compatible with existing approaches for handling noisy correspondences. Through extensive experiments on four benchmark datasets, we have proven the consistent superiority of GSC over current state-of-the-art methods.
\end{itemize}
\section{Related Works}
\label{Section: Related Works}

\subsection{Cross-modal Retrieval}
Cross-modal retrieval~\citep{liu2022regularizing, huang2022mack, pan2023fine, fu2023learning}, which focuses on using information from one modality to query the most relevant data in other modalities, has been a key area of research in multi-modal learning~\citep{huang2022modality, bachmann2022multimae, ge2023improving, schlarmann2023adversarial}. Due to the restriction of large-scale annotated multi-modal corpus~\citep{johnson2019mimic, kiela2018efficient, marin2019recipe1m+}, the unsupervised guided framework that directly aligns representations across modalities has always been the mainstream. SGRAF~\citep{diao2021similarity} introduces Graph Neural Network to establish graph correspondences and an attention mechanism to select the most representative alignments, enhancing the precision of cross-modal similarity assessments. However, these cross-modal retrieval methods highly rely on well-aligned data while amending the existence of noisy correspondence. 

% \vskip -0.2in
\subsection{Noisy Correspondence Learning}
% NCR DECL BICRO MSCN
Noisy correspondence~(NC) is first aroused in~\citep{huang2021learning}, which is a novel paradigm in the field of noise learning~\citep{hu2021learning, yao2018deep, yao2019safeguarded, yao2023latent}, referring to the mismatched pairs within the multi-modal dataset. 
To tackle this problem, NCR~\citep{huang2021learning} utilizes DivideMix~\citep{li2020dividemix} to distinguish clean pairs from noisy ones and rectify correspondence based on the memorization effect of DNNs. 
\citet{yang2023bicro} further improves NCR by switching to Beta Mixture Model and estimates soft correspondence labels by sample-wise comparison.
\citet{han2023noisy} proposes a meta-similarity correction network that reinterprets binary classification of correspondence as a meta-process, enhancing the process of data purification. 
Despite NCR-based models, other attempts like robust loss based methods have also been undertaken. \citet{qin2022deep} combines the idea of evidential learning with NC and puts forward a confidence-based method.
% , achieving better retrieval performance under extreme noise rates
\citet{chuang2022robust} introduces one effective robust symmetric contrastive loss~\citep{zhou2022contrastive, NEURIPS2023_40bb79c0, hong2023harmonizing}.
Although these models have showcased promising performances, they only cast their spotlight on interactions across modalities, which is insufficient in utilizing the semantic-abundant cross-modal data, further motivating us to take not only cross-modal but also intra-modal together into consideration to help mitigate NC.

\section{Proposed Method}
\label{Section: Method}
\subsection{Preliminary}

We start by defining notations for cross-modal retrieval in the presence of noisy correspondences, employing the widely studied image-text retrieval task as a generalized exemplar. Consider a multimodal dataset $\mathcal{D} = \{I_i, T_i, y_i\}_{i=1}^N$, where each $\{I_i, T_i\}$ represents the $i$-th image-text pair, standard retrieval models project these data pairs into a shared representation space using separate encoders $f$ for images and $g$ for texts. Then similarity scores between the representations are computed through cosine similarity or an inference model, denoting as $S(f(I), g(T))$, or $\langle I, T \rangle$ in brief. The associated label $y_i$ indicates whether the pair is positively correlated ($y_i=1$) or not ($y_i=0$). Note that these labels may contain noise, as pairs in the dataset are often presumed to be matched.

\subsection{Geometrical Structure Consistency Learning}
\label{Section: Motivation}

\begin{figure}[t]
\begin{center}
\includegraphics[width=0.47\textwidth]{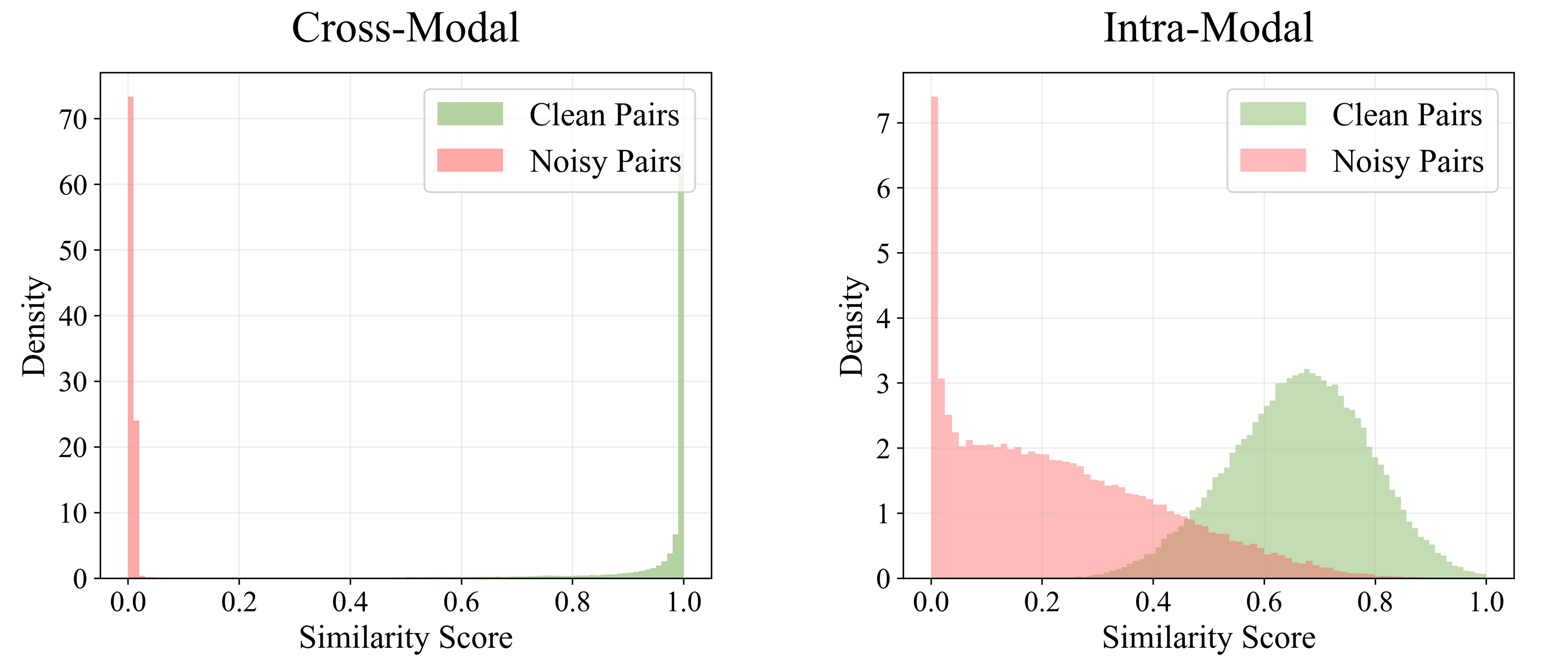}
% [width=1.30\textwidth]
\caption{Geometrical Structure Consistency helps discriminate samples with noisy correspondence. The model is first trained on clean Flickr30K dataset, then evaluated on the same dataset with 40\% simulated noise. \textbf{Left:} Calculated cross-modal similarity scores of both clean and noisy samples. \textbf{Right:} Calculated intra-modal similarity scores of both clean and noisy samples.}
\label{Figure: Motivation}
\end{center}
\vskip -0.3in

\end{figure}

\noindent
\textbf{Motivation.} The core concept of GSC is to preserve the consistency of geometrical structures and distinguish samples with noisy correspondence through structural differences. Initially, we demonstrate these differences through a straightforward experiment. As shown in Fig.~\ref{Figure: Motivation}, a retrieval model is optimized on a clean dataset to maintain consistent cross-modal and intra-modal structures, which is then assessed on a dataset with simulated noise. During the experiment, significant discrepancy between clean and noisy samples can be observed in both cross-modal and intra-modal structures. Specifically, for cross-modal structure, clean samples typically possess higher similarity scores than those noisy ones, exhibiting a disparity in distribution. For intra-modal structure, the calculated similarities between intra-modal structures of clean and noisy samples manifest a bimodal distribution with most values of noisy samples lower than 0.5 threshold, suggesting noisy samples tend to have asymmetric intra-modal structures.

\noindent
\textbf{Geometrical Structure Consistency.} Here, we give definitions to both cross-modal and intra-modal geometrical structures and the corresponding training objectives. From the cross-modal aspect, the geometrical structure is defined by the similarities between representations across different modalities. Considering an example of a given query image $I_i$, the cross-modal geometrical structure can be represented as $\mathcal{G}_{\text{CM}}^i = \{\langle I_i, T_j\rangle\}_{j=1}^N$. GSC preserves the consistency of this structure by minimizing the expected risk for cross-modal objective, as expressed in the following equation,
\begin{equation}
\mathcal{R}_{\mathcal{L}_{\text{CM}}}(f, g) = \min_{f, g} \mathbb{E}_{(I, T, y) \sim \mathcal{D}} \left[ \mathcal{L}_{\text{CM}}(\langle I, T\rangle, y) \right]
\label{Equal: Cross-modal Loss}
\end{equation}

\noindent
where $\mathcal{L}_{\text{CM}}$ is the cross-modal loss function, typically a contrastive or triplet loss in line with conventional retrieval models. The goal is to align the cross-modal representations according to the correspondence label $y$, thus the similarity between matching data pairs can be maximized, contrasting to other data pairs.

From the intra-modal aspect, the geometrical structure refers to the similarities within the modality. The intra-model structure for the $i$-th sample then denotes as $\mathcal{G}_{\text{IM}}^i = \{\langle I_i, I_j\rangle,\langle T_i, T_j\rangle \}_{j=1}^N$, where $\langle I_i, I_j\rangle$ and $\langle T_i, T_j\rangle$ represent the pairwise similarities among images and texts, respectively. To uphold the consistency of such structure, GSC incorporates the intra-modal objective as following,
\begin{equation}
 \mathcal{R}_{\mathcal{L}_{\text{IM}}}(f, g) = \min_{f, g} \mathbb{E}_{(I, T, y) \sim \mathcal{D}} \left[ \mathcal{L}_{\text{IM}}(\langle I, I\rangle, \langle T, T \rangle, y) \right]
\label{Equal: Intra-modal Loss}
\end{equation}

\begin{figure*}[!t]
\begin{center}
\includegraphics[width=1\textwidth]{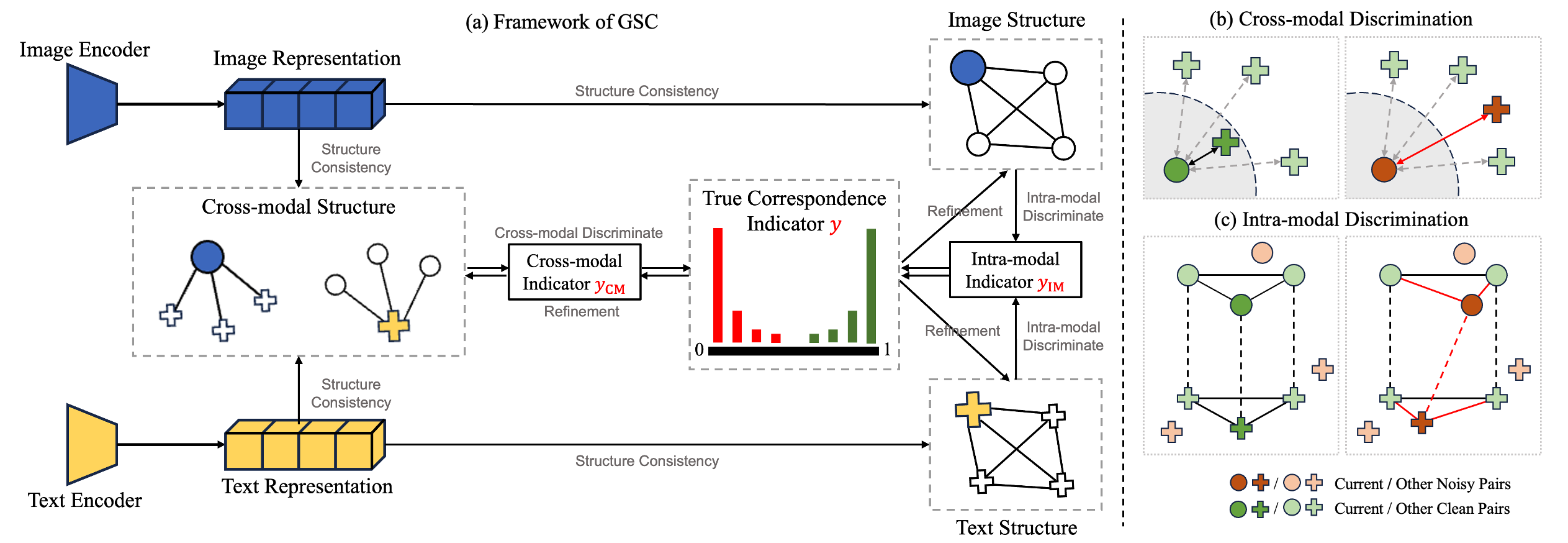}
% [width=1.30\textwidth]
\caption{An overview of GSC. \textbf{Left:} The framework of GSC. GSC first extracts image and text representations through separate encoders, then simultaneously optimizes cross-modal and intra-modal objectives to preserve geometrical structure consistency. GSC leverages both structures to discriminate noisy samples and estimate the true correspondence indicator $y$, which can be further utilized to purify the overall learning.
% GSC simultaneously optimizes both cross-modal and intra-modal objectives to preserve geometrical structure consistency and discriminate samples with noisy correspondence, while the estimated correspondence labels can be further utilized to refine both objective learning. 
\textbf{Right:} GSC discriminates noisy samples by structural differences from both cross-modal and intra-modal aspects.}
\label{Figure: Structure}
\end{center}
\vskip -0.3in
\end{figure*}

\noindent
In this equation, $\mathcal{L}_{\text{IM}}$ denotes the intra-modal loss function. The intra-modal objective ensures that the intra-modal structures of matching samples are constrained to be similar across modalities. Notably, optimizing without maintaining intra-modal structure is considered sub-optimal for inconsistent reasoning~\citep{goel2022cyclip, jiang2023understanding}. Thus the introduction of intra-modal structure consistency can also directly benefit multimodal representation learning. As illustrated in Fig.~\ref{Figure: Structure}(a), GSC simultaneously optimizes both objectives to establish stable cross-modal and intra-modal structures.

\subsection{Noise Discrimination \& Purification}

Deep neural networks typically exhibit the memorization effect that tends to initially learn the clean patterns within the dataset before over-fitting on noise. Leveraging this, GSC is able to learn a well-established structure in the early stage, which can be further utilized to predict accurate correspondence indicator.

\noindent
\textbf{Cross-modal Discrimination.} As illustrated in Fig.~\ref{Figure: Structure}(b), based on the well-established cross-modal structure in the early stage, clean data pairs are expected to exhibit more closely aligned cross-modal representations compared to noisy pairs. GSC leverages this structural discrepancy and introduces a function to signify a cross-modal bidirectional correspondence indicator.
\begin{equation}
\small
 y_{\text{CM}}^i = \frac{1}{2}\big[\frac{\exp{(\big<I_i, T_i \big>/\tau_1)}}{\sum_{j=1}^N\exp{(\big<I_i, T_j\big>/\tau_1)}} + \frac{\exp{(\big<I_i, T_i \big>/\tau_1)}}{\sum_{j=1}^N\exp{(\big<I_j, T_i\big>/\tau_1})}\big]
\label{Equal: Cross-modal Indicator}
\end{equation}

\noindent
where $\tau_1$ is the temperature coefficient. Take the former half as an example, it measures the proportion of similarity between the current data pairs $I_i$ and $T_i$ against the summation of similarity between the $I_i$ and all text data. For a clean sample, the similarity between current $I_i$ and $T_i$ should dominate the proportion, thus the value of indicator should approach 1. Conversely, for a noisy sample, the similarity between current $I_i$ and $T_i$ would be close to 0, resulting in the indicator's value trending toward 0.

\noindent
\textbf{Intra-modal Discrimination.} For the intra-modal aspect, as illustrated in Fig.~\ref{Figure: Structure}(c), matched image-text pairs should have similar intra-modal structures that mirror each other, whereas mismatched pairs would possess distinct structures that reflect their divergent positions. Specifically, we employ cosine similarity to measure the resemblance between intra-modal structures as $\mathcal{S}_{\text{IM}}^i = \cos(\{\langle I_i, I_j\rangle, \langle T_i, T_j\rangle\}_{j=1}^N)$. During experiments, the cosine similarity scores of clean samples are observed consistently higher than those of noisy samples during experiments, presenting a bimodal distribution of scores across the dataset (demonstrated in Fig.~\ref{Figure: Ablation_four_figures}(b)). This distribution can be accurately modeled using a two-component Gaussian Mixture Model (GMM)~\citep{li2020dividemix}, described by the equation:
\begin{equation}
\small
 p(\mathcal{S}_{\text{IM}}) = \sum_{k=1}^K\alpha_k \phi(\mathcal{S}_{\text{IM}}|k),\  y_{\text{IM}}^i = \frac{\alpha_{k_l}\phi(\mathcal{S}_{\text{IM}}^i|k_l)}{\sum_{k=1}^K \alpha_k\phi(\mathcal{S}_{\text{IM}}^i|k)}
 \label{Equal: GMM}
\end{equation}

\noindent
Specifically, $\alpha_k$ denotes the $k$-th coefficient and $\phi(\textbf{s}_{\text{IM}}|k)$ is the probability density for that component. The second equation is the estimation of the intra-modal correspondence indicator $y^i_{\text{IM}}$. It calculates the probability of an observed sample belonging to the cleaner component, denoted by $k = k_l$. This probability approaches 1 for clean samples and 0 for noisy samples, thereby enabling the distinction of samples with noisy correspondence.

So far, we have estimated the true correspondence labels, $y_{\text{CM}}$ and $y_{\text{IM}}$, by leveraging the structural differences in both cross-modal and intra-modal contexts. Our objective is to optimally utilize these two labels to surmount the respective challenges and accurately identify all samples with noisy correspondence. Therefore, we define the final correspondence label for each sample as the minimum of the two labels, which can expressed as below,
\begin{equation}
y^i = \min\{y_{\text{CM}}^i, y_{\text{IM}}^i\}
\end{equation}

\noindent
\textbf{Noise purification.} We address the issue of noisy correspondence by refining both cross-modal and intra-modal objectives. Since the estimated label is a soft label with values in the range of [0, 1] which can directly reflect the degree of true correspondence, we can seamlessly apply it to the loss functions on a sample-wise basis. For the cross-modal objective, we choose the widely-applied contrastive loss as the loss function. The purified cross-modal loss can be denoted as follows,
\begin{equation}
\begin{aligned}
 \mathcal{L}_{\text{CM}} =& -\frac{1}{2N}\sum_{i=1}^Ny^i\log{\frac{\exp{(\big<I_i, T_i\big>/\tau_1)}}{\sum_{j=1}^N\exp{(\big<I_i, T_j\big>/\tau_1)}}} \\
 & -\frac{1}{2N}\sum_{j=1}^Ny^j\log{\frac{\exp{(\big<I_j, T_j\big>/\tau_1)}}{\sum_{i=1}^N\exp{(\big<I_i, T_j\big>/\tau_1)}}}
\label{Equal: Purified Contrastive Loss CM}
\end{aligned}
\end{equation}

\noindent
where $y$ is directly applied before the sample-wise loss. For the intra-modal side, in addition to sample-wise purification for the intra-modal loss, it is crucial to prevent the impact of noisy correspondences from distorting the calculation of the geometrical structure. Specifically, the distances from the queried sample to those samples with noisy correspondences should be excluded. The purified intra-model loss denotes as, 
\begin{equation}
\small
 \mathcal{L}_{\text{IM}} = -\frac{1}{N}\sum_{i=1}^N\log{\frac{\exp({\sum_{k=1}^N{y^k}\big<I_i,I_k\big>y^k\big<T_i,T_k\big>/\tau_2)}}{\sum_{j=1}^N\exp{(\sum_{k=1}^N{y^k}\big<I_i, I_k\big>y^k\big<T_j, T_k\big>/\tau_2)}}}
\label{Equal: Purified Contrastive Loss IM}
\end{equation}

\noindent
where $\tau_2$ is also a temperature coefficient and $y^k$ is multiplied to both $\langle I_i, I_k\rangle$ and $\langle T_i, T_k\rangle$ to precisely filter out the noise. Furthermore, noisy correspondences can also interfere with the discrimination of noisy samples during the computation of cosine similarity scores for intra-modal discrimination. Similar modifications are employed as follows,
\begin{equation}
 \mathcal{S}_{\text{IM}}^i = \frac{\sum_{j=1}^{N} {y^j}\langle I_i, I_j \rangle y^j\langle T_i, T_j \rangle}{\sqrt{\sum_{j=1}^{N} (y^j\langle I_i, I_j \rangle)^2} \sqrt{\sum_{j=1}^{N} (y^j\langle T_i, T_j \rangle)^2}}
 \label{Equal: Purified Cosine Similarity}
\end{equation}

\noindent
where $y^j$ is similarly multiplied. Considering the high computational cost for the entire dataset, we employ a Monte Carlo sampling approach to relax size $N$ to size $B$ of a mini-batch. The overall loss function of GSC is formulated as a weighted sum of two loss functions, as expressed in the following equation,
\begin{equation}
\label{Equal: Total Loss}
 \mathcal{L} = \mathcal{L}_{\text{CM}} + \gamma\mathcal{L}_{\text{IM}}
\end{equation}

\noindent
where $\gamma$ is the hyper-parameter keeping the balance between two losses to reach the best optimization.

\begin{algorithm}[t]
\KwIn{Multi-modal dataset $\mathcal{D} = \{I_i, T_i, y_i\}_{i=1}^N$}
{
Initialize parameters for networks $A$ and $B$ separately

\For{each epoch $t = 1,2,\ldots,T$}{
\For{network $k = A, B$}{
\For{each minibatch $\mathcal{B}$ from $\mathcal{D}$}{

            Compute modality representations: $f(I)$, $g(T)$
            
            % Calculate \textit{cross-modal} and \textit{intra-modal} similarity vectors $\textbf{s}_{\text{CM}}$ and $\textbf{s}_{\text{IM}}$
            % using Eq.~\ref{Equal: Cosine Similarity} and Eq.~\ref{zz}

            % Calculate \textit{intra-modal} similarity vector $\mathcal{S}_{\text{IM}}$

            Estimate \textit{cross-modal} indicator $y_{\text{CM}}[t]_k$ 
            by Eq.~\ref{Equal: Cross-modal Indicator}

            Estimate \textit{intra-modal} indicator $y_{\text{IM}}[t]_k$ 
            by Eq.~\ref{Equal: GMM}

            % Update both sets of labels using temporal ensembling

            Calaulate \textit{cross-modal} loss by Eq.~\ref{Equal: Purified Contrastive Loss CM}

            Calculate \textit{intra-modal} loss by Eq.~\ref{Equal: Purified Contrastive Loss IM}
            
            Train Net$_k$ by optimizing the combination of two losses using Eq.~\ref{Equal: Total Loss}
            
            % Optimize $\theta$ using losses from $Net_B$
            % Optimize $Net_k$ using losses from another network
            % Optimize $\theta'$ using losses from $Net_A$

}

Update $y_{\text{CM}}[t]$ and $y_{\text{IM}}[t]$ by Eq.~\ref{Equal: Momentum Update}

Update the final $y[t]$ for the other network.

}

}         
% Refine both $Net_A$ and $Net_B$ using the final estimated labels

\KwOut{Refined networks Net$_A$, Net$_B$}

}
\caption{Pipeline of learning with our GSC method.}
\label{Algorithm}
\end{algorithm}
% \vskip -0.3in

\subsection{Training Schedule}

To integrate the estimation of true correspondence labels with the enhancement of cross-modal retrieval learning, we adopt the temporal ensembling technique, drawing inspiration from \citet{liu2020earlylearning}, to iteratively update the estimated correspondence labels. Specifically, both $y_{\text{CM}}$ and $y_{\text{IM}}$ are updated through a momentum-based combination of the estimates from the current epoch $t$ and the previous epoch $t-1$ before taking minimum as shown below, 
\begin{equation}
\begin{aligned}
 y_{\text{CM}}^i[t] &= \beta_1 y_{\text{CM}}^i[t] + (1-\beta_1) y_{\text{CM}}^i[t-1], \\
 y_{\text{IM}}^i[t] &= \beta_2 y_{\text{IM}}^i[t] + (1-\beta_2) y_{\text{IM}}^i[t-1]
 \label{Equal: Momentum Update}
 \end{aligned}
\end{equation}

\noindent
where $\beta_1$ and $\beta_2$ are separate momentum. The estimation of true correspondence labels can be further improved by utilizing two separate neural networks, where the true correspondence labels for each network are computed from the output of the other network. The ablation in Section~\ref{Section: Experiment} shows that both strategies can significantly improve the performance. In conclusion, the overall procedure of our Geometrical Structure Consistency (GSC) method is depicted in the Algorithm~\ref{Algorithm}.

\subsection{Discussion}

The improvement of our proposed GSC mainly comes from preserving the geometrical structure consistency and better optimizing strategies, which is compatible with most existing methods. In terms of computational complexity, GSC does not introduce additional computational costs when integrated with a backbone that calculates similarity scores directly from representations, while it necessitates two extra forward passes when the backbone computes similarity scores using a similarity module. This requirement is significantly less demanding compared to MSCN~\citep{han2023noisy}, which involves computations for an additional meta-learning model, or BiCro~\citep{yang2023bicro}, which compares each sample against a clean subset. While RINCE~\citep{chuang2022robust} shares similar computational cost as GSC, the robust loss function without explicitly excluding noisy samples underperforms at higher noise levels. Furthermore, methods based on NCR framework, \textit{i.e.} NCR~\citep{huang2021learning}, MSCN and BiCro, rely on an inseparable dual-network structure, while GSC is effective with a single model. This attribute makes GSC more adaptable to larger models.

\begin{table*}[!t]
\caption{The retrieval performance on Flickr30K and MS-COCO datasets under 20\%, 40\% and 60\% noise rates separately. The best results and the second best results are respectively marked by \textbf{bold} and \underline{underline}.}
% The highest score is displayed in \textbf{bold}.}
\vskip -0.2in
\label{Table: MS-COCO and Flickr30K}
\begin{center}
\begin{small}
\setlength{\tabcolsep}{1.5mm}
\resizebox{2\columnwidth}{!}{%
\begin{tabular}{c | c | c c c c c c c | c c c c c c c}
\toprule
 & & \multicolumn{7}{c|}{Flickr30K} & \multicolumn{7}{c}{MS-COCO}\\

 Noise & Methods & \multicolumn{3}{c}{Image $\rightarrow$ Text} & \multicolumn{3}{c}{Text $\rightarrow$ Image} &  & \multicolumn{3}{c}{Image $\rightarrow$ Text} & \multicolumn{3}{c}{Text $\rightarrow$ Image}  & \\
 
 & & R@1 & R@5 & R@10 & R@1 & R@5 & R@10 & Sum & R@1 & R@5 & R@10 & R@1 & R@5 & R@10 & Sum\\

\midrule

 & SGR    & 55.9       & 81.5       & 88.9       & 40.2       & 66.8       & 75.3       & 408.6 & 25.7       & 58.8       & 75.1       & 23.5       & 58.9       & 75.1       & 317.1 \\
 & SGRAF       & 72.8       & 90.8       & 95.4       & 56.4       & 82.1       & 88.6       & 486.1 & 75.4       & 95.2       & 97.9       & 60.1       & 88.5       & 94.8       & 511.9 \\
 & NCR-SGR     & 73.5       & 93.2       & 96.6       & 56.9       & 82.4       & 88.5       & 491.1 & 76.6       & 95.6       & 98.2       & 60.8       & 88.8       & 95.0       & 515.0 \\
  20\%
 & DECL-SGRAF  & 77.5       & 93.8       & 97.0       & 56.1       & 81.8       & 88.5       & 494.7 & 77.5       & 95.9       & 98.4       & 61.7       & 89.3       & 95.4       & 518.2 \\
 & RINCE-SGR   & 72.1       & 92.2       & 95.7       & 54.9       & 79.8       & 85.3       & 480.0 & 73.8       & 95.6       & 98.5       & 61.7       & 89.2       & 94.7       & 513.5 \\
 & MSCN-SGR    & 77.4       & \textbf{94.9}       & \underline{97.6}       & 59.6       & 83.2       & 89.2       & 501.9 & 78.1       & \textbf{97.2}       & \underline{98.8}       & \underline{64.3}       & \underline{90.4}       & \underline{95.8}       & \underline{524.6} \\
 & BiCro-SGRAF & \underline{78.1}       & 94.4       & 97.5       & \textbf{60.4}       & \underline{84.4}       & \underline{89.9}       & \underline{504.7} & \underline{78.8}       & 96.1       & 98.6       & 63.7       & 90.3       & 95.7       & 523.2 \\
 % & GSC-SGR    & \textbf{78.3}	& \underline{94.6} & \textbf{97.8} & \underline{60.1} & \textbf{84.5} & \textbf{90.5} & \textbf{505.8} & 79.5       & 96.1       & 98.6       & 64.0        & 90.3       & 95.8       & 524.4 \\

  & GSC-SGR    & \textbf{78.3}	& \underline{94.6} & \textbf{97.8} & \underline{60.1} & \textbf{84.5} & \textbf{90.5} & \textbf{505.8} & \textbf{79.5}       & \underline{96.4}       & \textbf{98.9}       & \textbf{64.4}        & \textbf{90.6}       & \textbf{95.9}       & \textbf{525.7} \\

\midrule
 
 & SGR    & 4.1        & 16.6       & 24.1       & 4.1        & 13.2       & 19.7       & 81.8  & 1.3        & 3.7        & 6.3        & 0.5        & 2.5        & 4.1        & 18.4  \\
 & SGRAF       & 8.3        & 18.1       & 31.4       & 5.3        & 16.7       & 21.3       & 101.1 & 15.8       & 23.4       & 54.6       & 17.8       & 43.6       & 54.1       & 209.3 \\
 & NCR-SGR     & 68.1       & 89.6       & 94.8       & 51.4       & 78.4       & 84.8       & 467.1 & 74.7       & 94.6       & 98.0       & 59.6       & 88.1       & 94.7       & 509.7 \\
 40\%
 & DECL-SGRAF  & 72.7       & 92.3       & 95.4       & 53.4       & 79.4       & 86.4       & 479.6 & 75.6       & 95.5       & \textbf{98.3}       & 59.5       & 88.3       & 94.8       & 512.0 \\
 & RINCE-SGR   & 71.2       & 90.7       & 95.6       & 52.7       & 78.5       & 85.6       & 474.3 & 71.2       & 95.8       & 97.9       & 59.1       & 88.6       & 94.3       & 506.9 \\
 & MSCN-SGR    & 71.6       & \underline{92.8}       & \underline{96.2}       & 54.8       & 80.7       & \underline{87.4}       & 483.5 & 75.3       & 95.4       & 98.2       & 60.3       & 88.6       & 94.8       & 512.6 \\
 & BiCro-SGRAF & \underline{74.6}       & 92.7       & \underline{96.2}       & \underline{55.5}       & \underline{81.1}       & \underline{87.4}       & \underline{487.5} & \underline{77.0}       & \textbf{95.9}      & \textbf{98.3}       & \underline{61.8}       & \underline{89.2}       & \underline{94.9}       & \underline{517.1} \\
 & GSC-SGR     & \textbf{76.5}       & \textbf{94.1}       & \textbf{97.6}       & \textbf{57.5}       &\textbf{82.7}       & \textbf{88.9}       & \textbf{497.3} & \textbf{78.2}       & \textbf{95.9}       & 98.2       & \textbf{62.5}       & \textbf{89.7}       & \textbf{95.4}       & \textbf{519.9} \\

\midrule
 
 & SGR    & 1.5        & 6.6        & 9.6        & 0.3        & 2.3        & 4.2        & 24.5  & 0.1        & 0.6        & 1.0        & 0.1        & 0.5        & 1.1        & 3.4   \\
 & SGRAF       & 2.3        & 5.8        & 10.9       & 1.9        & 6.1        & 8.2        & 35.2  & 0.2        & 3.6        & 7.9        & 1.5        & 5.9        & 12.6       & 31.7  \\
 & NCR-SGR     & 13.9       & 37.7       & 50.5       & 11.0       & 30.1       & 41.4       & 184.6 & 0.1        & 0.3        & 0.4        & 0.1        & 0.5        & 1.0        & 2.4   \\
 60\%
 & DECL-SGRAF  & 65.2       & 88.4       & 94.0       & 46.8       & 74.0       & 82.2       & 450.6 & 73.0       & 94.2       & \underline{97.9}       & 57.0       & 86.6       & 93.8       & 502.5 \\
& RINCE-SGR   & 64.5       & 86.8       & 92.9       & 46.5       & 72.8       & 79.7       & 443.2 & 72.3       & 94.0       & \underline{97.9}       & \underline{58.4}       & 86.6       & 92.5       & 501.7 \\
 & MSCN-SGR    & \underline{68.8}       & 90.3       & \underline{94.4}       & 50.8       & 77.4       & 84.4       & 466.1 & 72.5       & 93.6       & 97.1       & 57.7       & 87.0       & \underline{93.9}       & 501.8 \\
& BiCro-SGRAF & 67.6       & \underline{90.8}       & \underline{94.4}       & \underline{51.2}       & \underline{77.6}       & \underline{84.7}       & \underline{466.3} & \underline{73.9}       & \underline{94.4}       & 97.8       & 58.3       & \underline{87.2}       & \underline{93.9}       & \underline{505.5} \\
 & GSC-SGR     & \textbf{70.8}       & \textbf{91.1}       & \textbf{95.9}       & \textbf{53.6}       & \textbf{79.8}       & \textbf{86.8}       & \textbf{478.0} & \textbf{75.6}       & \textbf{95.1}       & \textbf{98.0}        & \textbf{60.0}        & \textbf{88.3}       & \textbf{94.6}       & \textbf{511.7} \\
\bottomrule
\end{tabular}}
\end{small}
\end{center}
\vskip -0.3in
\end{table*}

\begin{table}[h]
\caption{Experimental results on NUS-WIDE dataset. Only results of best SOTAs are presented. Best results are marked by \textbf{bold}.}
\vskip -0.2in

\label{Table: NUS-WIDE}
\begin{center}
\begin{small}
\setlength{\tabcolsep}{0.9mm}

\begin{tabular}{c | c | c c c c c c c}
\toprule

Noise & Method & \multicolumn{3}{c}{Image $\rightarrow$ Text} & \multicolumn{3}{c}{Text $\rightarrow$ Image}\\
& & R@1 & R@5 & R@10 & R@1 & R@5 & R@10 & Sum \\
\midrule
\multirow{3}{*}{\text{clean}}  & SGR  & 39.3          & 66.5          & 74.0          & 37.1          & 64.2          & 73.6          & 354.7          \\
                      & NCR & 40.0          & \textbf{67.3}          & \textbf{74.6}          & 38.9          & 65.9          & 74.3          & 361.0          \\
                      & GSC  & \textbf{42.9} & 65.9 & 74.0 & \textbf{39.8} & \textbf{66.0} & \textbf{74.5} & \textbf{363.1} \\
\midrule
\multirow{3}{*}{20\%} & SGR  & 33.4          & 60.7          & 69.8          & 31.1          & 58.7          & 68.2          & 321.9          \\
                      & DECL & 36.1          & 63.0          & 72.5          & 36.4          & 60.6          & 72.0          & 340.6          \\
                      & GSC  & \textbf{37.9} & \textbf{65.2} & \textbf{73.4} & \textbf{38.3} & \textbf{63.8} & \textbf{73.1} & \textbf{351.7} \\
\midrule
\multirow{3}{*}{40\%} & SGR  & 30.8          & 57.9          & 67.3          & 28.4          & 55.9          & 65.5          & 305.8          \\
                      & DECL & 34.1          & 59.5          & 68.8          & 34.5          & 60.0          & 69.1          & 326.0          \\
                      & GSC  & \textbf{37.0} & \textbf{63.5} & \textbf{71.3} & \textbf{36.8} & \textbf{62.4} & \textbf{71.1} & \textbf{342.1} \\
\midrule
\multirow{3}{*}{60\%} & SGR  & 25.0          & 50.2          & 59.8          & 25.0          & 49.3          & 60.3          & 269.6          \\
                      & BiCro & 27.8          & 53.4          & 63.2          & 27.9          & 52.2          & 64.1          & 288.6          \\
                      & GSC  & \textbf{31.6} & \textbf{57.4} & \textbf{66.1} & \textbf{31.1} & \textbf{57.8} & \textbf{66.3} & \textbf{310.3}\\
\bottomrule
\end{tabular}
\end{small}
\end{center}
\vskip -0.3in
\end{table}

\section{Experiment}
\label{Section: Experiment}

\subsection{Datasets and Evaluation Metrics}
\noindent
\textbf{Datasets.} Following the experimental settings and dataset splits in \citet{huang2021learning}, four widely-used image-text retrieval datasets are introduced to evaluate our method:

\begin{itemize}
    \item Flickr30K~\citep{young2014image} contains 31,000 images with five captions each, collected from the Flickr website. We assign 1,000 image-text pairs for validation, 1,000 image-text pairs for testing and the rest for training.
    \item MS-COCO~\citep{lin2014microsoft} includes 123,287 images with five captions each. We assign 5,000 image-text pairs for validation, 5,000 image-text pairs for testing and the rest for training. MS-COCO can be either evaluated using whole 5,000 test set or average of 5-fold 1,000 test sets~\citep{karpathy2015deep}.
    \item NUS-WIDE~\cite{chua2009nus} includes 269,648 image-text pairs collected from Flickr, in which we randomly sample 1,000 image-text pairs for validation, 1,000 image-text pairs for testing and 100,000 image-text pairs for training.
    \item Conceptual Captions~\citep{sharma2018conceptual} is a large-scale dataset  automatically harvested from the Internet, therefore about 3\%-20\% image-text pairs in the dataset are mismatched or weakly-matched~\citep{sharma2018conceptual}. We use a subset named CC152K in our experiments, in which we assign 1,000 image-text pairs for validation, 1,000 image-text pairs for testing and 150,000 image-text pairs for training.
\end{itemize}

\noindent
\textbf{Evaluation Protocol.} We evaluate the retrieval performance with the recall rate at K (R@K) metric. In a nutshell, 
% R@K is the proportion for the relevant items retrieved in the closest K items to the query. 
R@K measures the proportion of relevant items retrieved within the top K items closest to the query. 
In our experiments, we take image and text as queries, respectively, and report R@1, R@5, R@10 results and their sum for a comprehensive evaluation.

\begin{table*}[!h]
\begin{minipage}{0.47\textwidth}
\caption{The retrieval performance on CC152K dataset. The best results and the second best results are respectively marked by \textbf{bold} and \underline{underline}.}
\begin{center}
\label{Table: CC152K}
\begin{small}
\vskip -0.2in
\setlength{\tabcolsep}{0.85mm}

\begin{tabular}{c | c c c c c c c}

\toprule
 & \multicolumn{7}{c}{CC152K} \\

 Methods & \multicolumn{3}{c}{Image $\rightarrow$ Text} & \multicolumn{3}{c}{Text $\rightarrow$ Image} &  \\
 
 & R@1 & R@5 & R@10 & R@1 & R@5 & R@10 & Sum\\

\midrule

SGR         & 11.3       & 29.7       & 39.6       & 13.1       & 30.1       & 41.6       & 165.4 \\
SGRAF       & 32.5       & 59.5       & 70.0       & 32.5       & 60.7       & 68.7       & 323.9 \\
NCR-SGR     & 39.5       & 64.5       & 73.5       & 40.3       & 64.6       & 73.2       & 355.6 \\
DECL-SGRAF  & 39.0       & 66.1       & 75.5       & 40.7       & 66.3       & \underline{76.7}       & 364.3 \\
RINCE-SGR   & 35.9       & 63.0       & 73.8       & 37.6       & 65.0       & 73.4       & 348.7 \\
MSCN-SGR    & 40.1       & 65.7       & \underline{76.6}       & 40.6       & 67.4       & 76.3       & 366.7 \\
BiCro-SGRAF & \underline{40.8}       & \underline{67.2}       & 76.1       & \underline{42.1}       & \textbf{67.6}       & 76.4       & \underline{370.2} \\
GSC-SGR     & \textbf{42.1}       & \textbf{68.4}       & \textbf{77.7}       & \textbf{42.2}       & \textbf{67.6}       & \textbf{77.1}       & \textbf{375.1} \\

\bottomrule
\end{tabular}
\end{small}
\end{center}
\end{minipage}
\hfill
\begin{minipage}{0.48\textwidth}
\caption{Comparison with CLIP and NCR on MS-COCO 5K. CLIP-L and CLIP-B are abbreviations for CLIP (ViT-L/14) and CLIP (ViT-B/32). The best results are marked by \textbf{bold}.}
\vskip -0.2in
\label{tab_clip}
\begin{center}
\begin{small}
\setlength{\tabcolsep}{0.65mm}
\begin{tabular}{c|c|ccccccc}

\toprule

\multirow{2}{*}{Noise} & \multirow{2}{*}{Methods} & \multicolumn{3}{c}{\centering Image $\rightarrow$ Text}&\multicolumn{3}{c}{\centering Text $\rightarrow$ Image}\\
    
 & & R@1 & R@5 & R@10 & R@1 & R@5 & R@10 & Sum \\
\midrule
0\% & CLIP-L & 58.4 & 81.5 & 88.1 & 37.8 & 62.4 & 72.2 & 400.4\\
Zeroshot& CLIP-B & 50.2 & 74.6 & 83.6 & 30.4 & 56.0 & 66.8 & 361.6\\

\midrule
20\% & CLIP-B & 21.4 & 49.6 & 63.3 & 14.8 & 37.6 & 49.6 & 236.3\\
Finetune & NCR & 56.9 & 83.6 & 91.0 & 40.6 & 69.8 & 80.1 & 422.0\\
& GSC & \textbf{58.9} & \textbf{84.9} & \textbf{91.7} & \textbf{42.0} & \textbf{71.4} & \textbf{81.8} & \textbf{430.8}\\
\midrule
50\% & CLIP-B & 10.9 & 27.8 & 38.3 & 7.8 & 19.5 & 26.8 & 131.1\\
Finetune & NCR & 53.1 & 80.7 & 88.5 & 37.9 & 66.6 & 77.8 & 404.6\\
% & GSC & \textbf{53.8} & \textbf{81.1} & \textbf{88.9} & \textbf{39.1} & \textbf{67.9} & \textbf{78.9} & \textbf{409.6} \\
& GSC & \textbf{55.5} & \textbf{81.8} & \textbf{90.1} & \textbf{40.0} & \textbf{69.1} & \textbf{79.7} & \textbf{416.3} \\
\bottomrule
\end{tabular}
\end{small}
\end{center}
\end{minipage}
\vskip -0.2in
\end{table*}

\subsection{Implementation Details}
For all experiments, we apply the Adam optimizer~\citep{kingma2014adam} with the initial learning rate of $2 \times 10^{-4}$ which decays by $0.2$ in $15$ epochs. We train the model on one NVIDIA A100 GPU and select the model that performs best on the validation set for testing. The dimension of the common representation is set to $1024$.
% Following~\citep{guerrero2021}, we use ResNet50~\citep{he2016deep} pre-trained on ImageNet~\citep{deng2009imagenet} as the visual encoder, and use WordPiece~\citep{Wu2016GooglesNM} to transform each word into a vector of 768 dimensions followed by a 2-layer 2-head transformer encoder. 
For experiments besides ablation study, we set the batch size $B$ as $128$. The two temperature coefficients $\tau_1$ and $\tau_2$ are set to $0.07$ and $1$ in default. The hyperparameter $\lambda$ serving as the balancing ratio between $\mathcal{L}_{\text{CM}}$ and $\mathcal{L}_{\text{IM}}$ is set to $0.01$, and the two momentum $\beta_{1}$ and $\beta_{2}$ are set to $0.7$ respectively.

\subsection{Comparison with the State-of-the-Arts}
We conducted extensive evaluations against seven contemporary state-of-the-art methods on four benchmark datasets to validate the effectiveness of our proposed GSC model. These comparisons include two baseline models, \textit{i.e.}, SGR and SGRAF~\citep{diao2021similarity}, and five robust learning methods designed to handle noisy correspondences, \textit{i.e.}, NCR~\citep{huang2021learning}, DECL~\cite{qin2022deep}, RINCE~\citep{chuang2022robust}, MSCN~\citep{han2023noisy} and BiCro~\citep{yang2023bicro}. Notably, both DECL and BiCro are built upon a stronger SGRAF backbone, which is an ensemble of SGR and SAF. To thoroughly assess the robustness of GSC, we simulate various levels of noisy correspondences, namely 20\%, 40\%, and 60\%, by randomly shuffling the captions on MSCOCO and Flickr30K like~\citep{huang2021learning}. In addition, we extend our experiments to real-world noisy conditions using the CC152K dataset. Comprehensive comparison results are detailed in the supplementary material for fully demonstration of GSC.

\noindent
\textbf{Results on Flickr30K, MS-COCO and NUS-WIDE.} To evaluate the robustness of all methods under different extents of noise, we quantify the noise rate to 20\%, 40\% and 60\% on well-annotated datasets, as recorded in Tab.~\ref{Table: MS-COCO and Flickr30K} and Tab.~\ref{Table: NUS-WIDE}. The results demonstrate that GSC significantly outperforms established noisy correspondence methods such as NCR, DECL, RINCE, MSCN, and BiCro, achieving an average increase in recall sum score of 7.5\% on Flickr30K, 3.4\% on MS-COCO and 12.75\% on NUS-WIDE to the second best results, which indicates the better robustness of GSC. Notably, GSC also excels over BiCro and DECL under various conditions, even though they are implemented on the enhanced SGRAF backbone. Moreover, GSC can carry about more enhancement at higher noise rates, especially under 60\% noise level, proving that our method remains stable and reliable even in severely noisy conditions.

% \begin{table}[!t]
% \centering
% \caption{Comparison with CLIP and NCR on MS-COCO 5K. CLIP-L and CLIP-B are abbreviations for CLIP (ViT-L/14) and CLIP (ViT-B/32). ZS and FT are short for \textit{zero-shot} and \textit{fine-tune} respectively. The best results are marked by \textbf{bold}.}
% \begin{small}
% \setlength{\tabcolsep}{0.6mm}
% \begin{tabular}{c|c|ccccccc}
% \toprule
% \multirow{2}{*}{Noise} & \multirow{2}{*}{Methods} & \multicolumn{3}{c}{\centering Image $\rightarrow$ Text}&\multicolumn{3}{c}{\centering Text $\rightarrow$ Image}\\
    
%  & & R@1 & R@5 & R@10 & R@1 & R@5 & R@10 & Sum \\
% \midrule
% 0\% & CLIP-L & 58.4 & 81.5 & 88.1 & 37.8 & 62.4 & 72.2 & 400.4\\
% Zeroshot& CLIP-B & 50.2 & 74.6 & 83.6 & 30.4 & 56.0 & 66.8 & 361.6\\

% \midrule
% 20\% & CLIP-B & 21.4 & 49.6 & 63.3 & 14.8 & 37.6 & 49.6 & 236.3\\
% Finetune & NCR & 56.9 & 83.6 & 91.0 & 40.6 & 69.8 & 80.1 & 422.0\\
% & GSC & \textbf{58.9} & \textbf{84.9} & \textbf{91.7} & \textbf{42.0} & \textbf{71.4} & \textbf{81.8} & \textbf{430.8}\\
% \midrule
% 50\% & CLIP-B & 10.9 & 27.8 & 38.3 & 7.8 & 19.5 & 26.8 & 131.1\\
% Finetune & NCR & 53.1 & 80.7 & 88.5 & 37.9 & 66.6 & 77.8 & 404.6\\
% & GSC & \\
% \bottomrule                  
% \end{tabular}
% \label{tab_clip}
% \end{small}
% \end{table}

\noindent
\textbf{Results on CC152K.} To further validate GSC in handling with noisy correspondence in real-world scenarios, we additionally conduct tests on CC152K dataset, detailed in Tab.~\ref{Table: CC152K}. According to the results, GSC achieves the best performance with an overall score of 375.1\%, surpassing the second best method BiCro by 4.9\%. Moreover, GSC brings about a larger gain of 209.7\% to its backbone SGR, which is significantly higher than the improvement of 46.3\% brought about by BiCro to its backbone SGRAF. The results affirm GSC's capability to manage not only simply simulated but also complex, real-world noisy correspondences.

% \begin{figure}[t!]
% \centerline{\includegraphics[width=0.8\linewidth]{pics/abl.pdf}}
% \caption{Ablation study on Flicker30K with 40\% noise with different hyper-parameter combinations. $\gamma$ is the balancing parameter between $\mathcal{L}_{\text{CM}}$ and $\mathcal{L}_{\text{IM}}$, and $\beta_1$ and $\beta_2$ are two momentums during correspondence purification updating of inter-modal and intra-modal respectively.}
% \label{fig:abl}
% \end{figure}

\begin{table}[!t]

\caption{Ablation study on Flicker30K with 40\% noise with different components in GSC. The best results are marked by \textbf{bold}.}
\vskip -0.2in
\label{Table: Ablation Study}
\begin{center}
\begin{small}
\setlength{\tabcolsep}{0.6mm}

\begin{tabular}{c c c | c c c c c c c}
\toprule
\multirow{2}{*}{Momen.}& \multirow{2}{*}{Dual}& \multirow{2}{*}{$\mathcal{L}_{\text{IM}}$} & \multicolumn{3}{c}{Image $\rightarrow$ Text} & \multicolumn{3}{c}{Text $\rightarrow$ Image}\\
 & & & R@1 & R@5 & R@10 & R@1 & R@5 & R@10 & Sum \\
\midrule
\checkmark & \checkmark &  & 72.3 & 92.8 & 96.5 & 56.5 & 82.1 & 88.8 & 489.0 \\ % ELCL+

\checkmark &  & \checkmark & 73.0 & 91.5 & 95.9 & 54.5 & 80.4 & 87.6 & 482.9 \\ % ELCLTOPO
 & \checkmark & \checkmark & 74.5 & 92.5 & 96.9 & 57.0 & 82.0 & 88.3 & 491.1 \\
\checkmark & \checkmark & \checkmark & \textbf{76.5} & \textbf{94.1} & \textbf{97.6} & \textbf{57.5}  & \textbf{82.7} & \textbf{88.9}  & \textbf{497.3}    \\
\bottomrule
\end{tabular}
\end{small}
\end{center}
\vskip -0.2in
\end{table}

\begin{figure}[h]
\centering
  \begin{minipage}{0.23\textwidth}
    \centering
    \includegraphics[width=\textwidth]{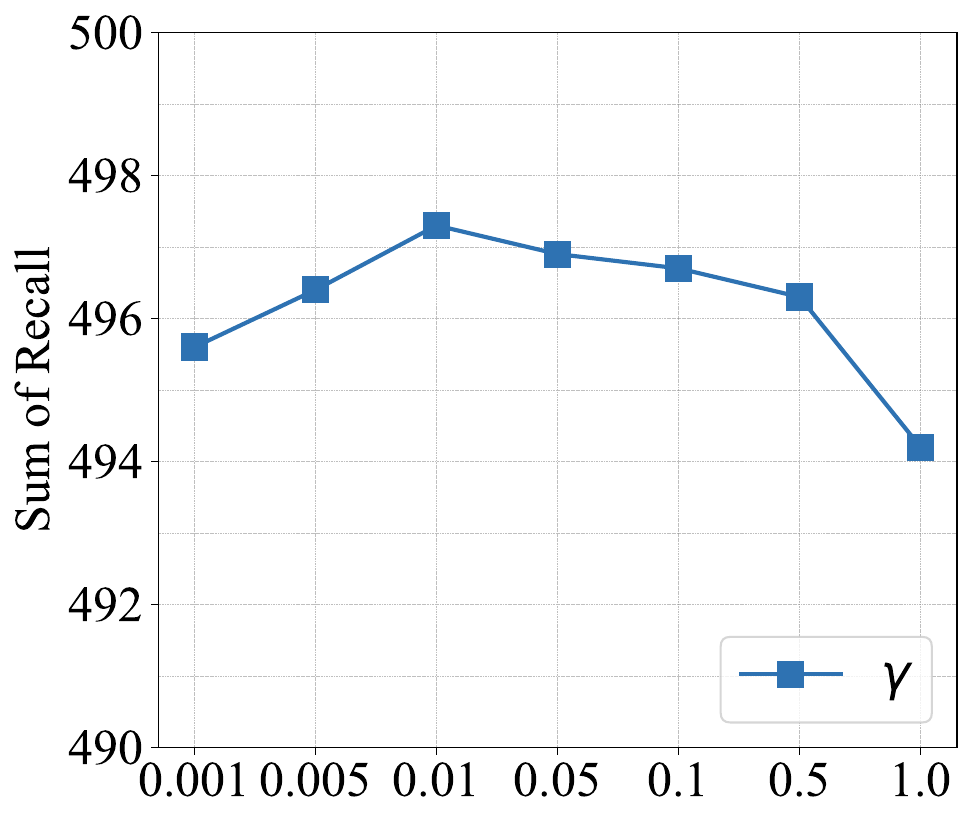}
  \end{minipage}
  \begin{minipage}{0.23\textwidth}
    \centering
    \includegraphics[width=\textwidth]{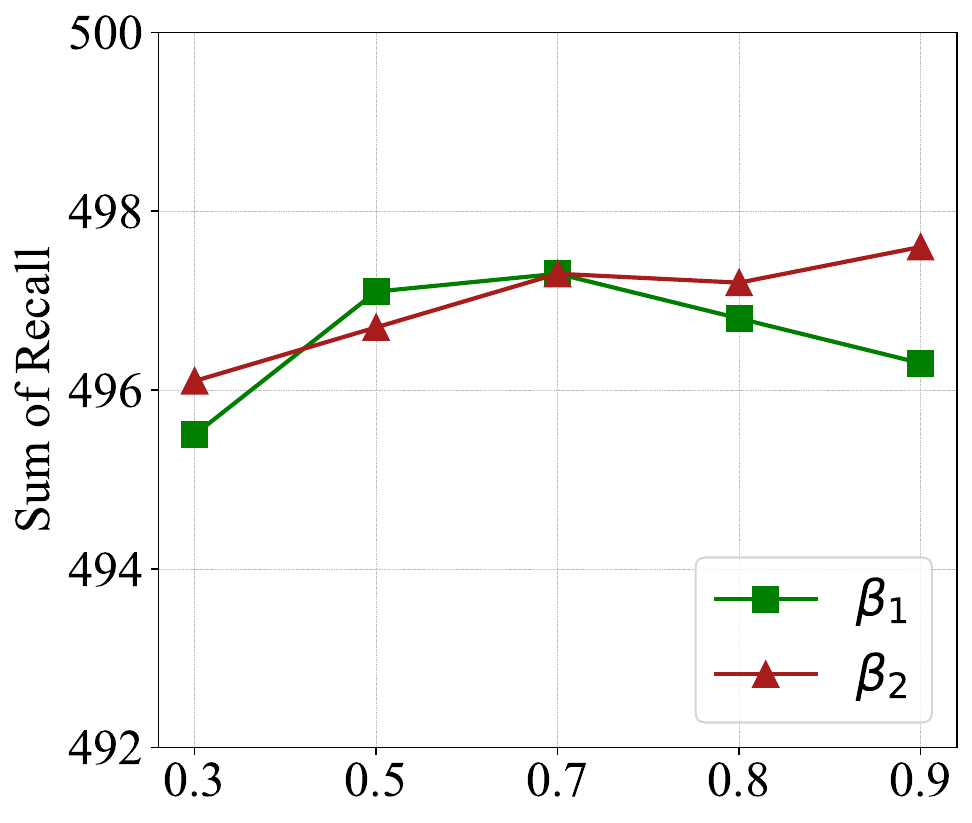}
  \end{minipage}

\caption{Analysis of different hyper-parameter combinations on Flicker30K with 40\% noise. \textbf{Left:} $\gamma$ is the balancing parameter between $\mathcal{L}_{\text{CM}}$ and $\mathcal{L}_{\text{IM}}$. \textbf{Right:} $\beta_1$ and $\beta_2$ are separate momentums for the cross-modal and intra-modal temporal ensembling.}
\label{Figure: Hyper-parameter}
\vskip -0.2in
\end{figure}

\begin{figure*}[!t]
% \vskip -0.13in
\begin{center}

\centerline{\includegraphics[width=1\textwidth]{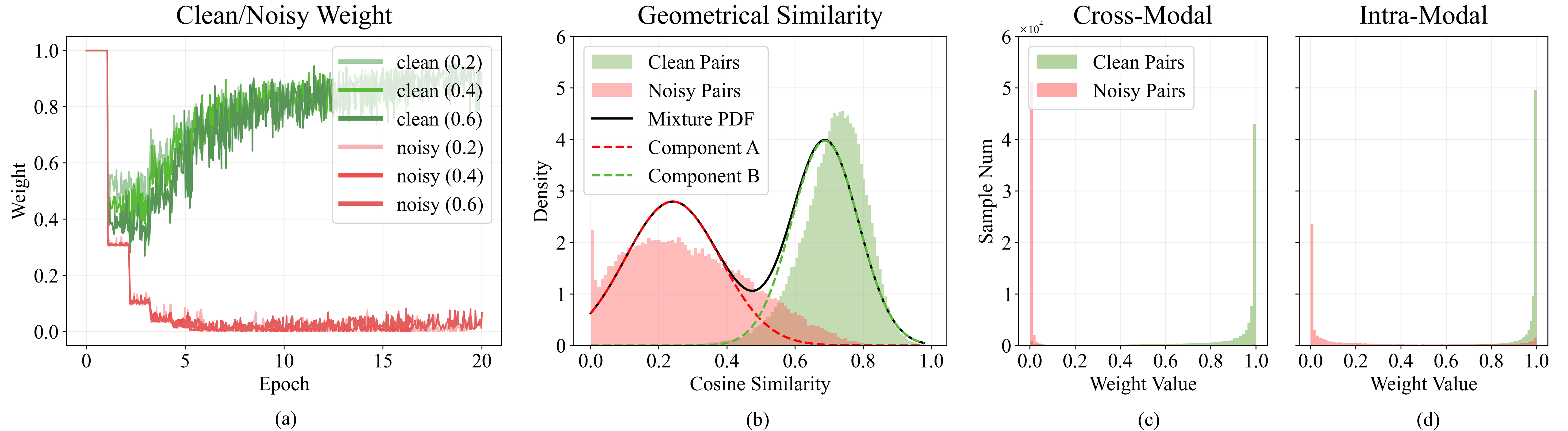}}
% \vskip -0.1in
% [width=1.30\textwidth]
\caption{
(a) The changing values of clean and noisy sample weight when the noise rate is 20\%, 40\%, and 60\%.
(b) Distribution of intra-modal geometrical similarity, including PDFs of clean and noisy pair similarities and estimated Gaussian distribution components. 
(c) Cross-modal weight distributions of GSC on clean and noisy pairs. 
(d) Intra-modal weight distributions of GSC on clean and noisy pairs. 
Experiments from (b) to (d) are conducted on Flickr30K with the noise rate of 0.4.}
\label{Figure: Ablation_four_figures}
\end{center}
\vskip -0.35in
\end{figure*}

\noindent
\textbf{Comparison to pre-trained model.} In line with \citet{huang2021learning}, we compare GSC to the pre-trained CLIP model~\citep{radford2021learning} on the MS-COCO dataset. CLIP is a well-known large pre-trained model trained on a massive 400 million image-text pair dataset harvested from the Internet, which can inevitably include samples with noisy correspondence. Here, we report the zero-shot and fine-tuning performances of different CLIP models, together with NCR. Results indicate a notable performance decline in CLIP models when fine-tuning with 20\% and 50\% noise levels. On the contrary, GSC not only withstands but excels over zero-shot CLIP under 50\% noise, emphasizing the importance of addressing data mismatches and the robustness of GSC.

\subsection{Experimental Analysis}

\noindent
\textbf{Ablation study.} We show the effect of each component of GSC in Tab.~\ref{Table: Ablation Study}. The ablation studies are conducted without temporal ensembling (Momen. in the table), dual networks or intra-modal learning separately. Specifically for the ablation study for temporal ensembling, we use a 5-epoch warm-up stage to replace the technique. According to the results, all components are important to achieve advantageous results. Notably, the performance of GSC with single model still outperforms most robust methods, including methods with dual networks like NCR and DECL, which further proves the effectiveness and high efficiency.

\begin{table}[!t]
\caption{Analysis of different batch sizes on Flicker30K with 40\% noise. The best results are marked by \textbf{bold}.}
\vskip -0.2in

\label{Table: Batch}
\begin{center}
\begin{small}
\setlength{\tabcolsep}{0.9mm}

\begin{tabular}{c | c c c c c c c}
\toprule
\multirow{2}{*}{Batch Size} & \multicolumn{3}{c}{Image $\rightarrow$ Text} & \multicolumn{3}{c}{Text $\rightarrow$ Image}\\
 & R@1 & R@5 & R@10 & R@1 & R@5 & R@10 & Sum \\
\midrule
32 & 73.6 & 92.1 & 95.4 & 54.4 & 81.0 & 87.4 & 483.9 \\
% 64 & 73.8 & 93.6 & 96.3 & 56.0 & 81.9 & 88.2 & 489.8\\
64 & 75.3 & 94.0 & 96.7 & 55.9 & 81.7 & 88.2 & 491.8 \\
128 & 76.5 & \textbf{94.1} & 97.6 & 57.5 & \textbf{82.7} & 88.9 & 497.3  \\
192 & \textbf{76.7} & 94.0 & \textbf{97.7} & \textbf{57.6} & \textbf{82.7} & \textbf{89.1} & \textbf{497.8}\\
\bottomrule
\end{tabular}
\end{small}
\end{center}
\vskip -0.3in
\end{table}

\noindent
\textbf{Impacts of hyper-parameters $\gamma$, $\beta_1$ and $\beta_2$.} The GSC method incorporates three main hyper-parameters including $\gamma$, $\beta_1$ and $\beta_2$ with their effects detailed in Fig.~\ref{Figure: Hyper-parameter}. $\gamma$ strikes a balance between cross-modal and intra-modal optimization. According to the results, GSC shows stability for $\gamma\in[0.005, 0.1]$, while $\gamma=0.01$ is chosen for optimal performance. $\beta_1$ and $\beta_2$ are the momentum coefficients ensuring steady updates for temporal ensembling. The results indicate stable performance with parameter values higher than 0.5. Lower beta values may laten timely updates, potentially causing the model to overfit on noise.

\noindent
\textbf{Impact of batch size.} We also explore the model performance under different batch sizes during training. As shown in Tab.~\ref{Table: Batch}, as the batch size increases, the retrieval performance of the model steadily improves. Specifically, when the batch size is increased from 32 to 128, there is a significant enhancement from 483.9\% to 497.3\% as to the sum of recall, and further expanding the batch size from 128 to 192 results in only marginal growth, which indicates that larger batch size helps in consolidating the stableness of model structure until reaching a proper point.

\noindent
\textbf{Experimental visualization.} To further offer insights of GSC against noisy correspondence, we visually present the value curves and weight distributions of predicted correspondence labels in Fig.~\ref{Figure: Ablation_four_figures}. Figure (a) shows the stability of predicted labels across varying noise levels, with minimal fluctuation for clean samples and consistently low values for noisy ones. Figure (b) depicts a bimodal distribution of intra-modal cosine similarities, which can be well-fitted by a two-component GMM. Figures (c) and (d) confirm that both cross-modal and intra-modal structures effectively distinguish noisy samples, with discrimination accuracy of approximately 0.96 for cross-modal and about 0.91 for intra-modal, culminating in an overall accuracy of about 0.98. Such visualization explains the reason for the steady and reliable performance of GSC under different noise rates.
\section{Conclusion}
% Noisy correspondences that refer to mismatches in cross-modal data pairs, are prevalent in human-annotated or web-crawled datasets. Prior approaches often adapted methods for unimodal noisy labels, overlooking the unique challenges of multimodal data. Actually, the existence of noisy correspondence can undermine cross-modal models by disrupting the geometrical structure of representations from both inter-modal and intra-modal aspects. To tackle this problem, we introduce the Geometrical Structure Consistency (GSC) method to infer the true correspondence. Specifically, GSC ensures the preservation of geometrical structures within and between modalities, allowing for the accurate discrimination of noisy samples based on structural differences. Utilizing these inferred true correspondence labels, GSC refines the learning of geometrical structures by filtering out the noisy samples.  Our experiments across three well-known cross-modal retrieval datasets confirm that our method effectively identifies noisy samples under various conditions of noisy correspondence, significantly outperforming the current leading methods.
In this paper, we propose Geometrical Structure Consistency (GSC) learning framework to mitigate the problem brought by noisy correspondence. Specifically, we identify the impact of noisy correspondence on both cross-modal and intra-modal geometrical structures. Leveraging the structural differences between noisy and clean pairs within a well-established structure, our approach infers accurate correspondence labels for each data pair. The inferred labels are further utilized to refine the consistent learning of structures. GSC can seamlessly integrate with most existing retrieval methods. Extensive experiments across various cross-modal benchmark datasets showcase the robustness and effectiveness of our proposed GSC method across diverse settings.

\small
\noindent
\textbf{Acknowledgement.} This work is supported by the National Key R\&D Program of China (No.2022ZD0160702), STCSM (No.22511106101, No.22511105700, No.21DZ1100100), 111 plan (No.BP0719010) and National Natural Science Foundation of China (No.62306178).

{
    \small
    \bibliographystyle{ieeenat_fullname}
    \bibliography{main}
}


\end{document}